\documentclass[11pt]{article}

\usepackage[margin=1in]{geometry}
\usepackage{setspace}
\usepackage{lineno}
\usepackage{amsmath}       

\usepackage{authblk}

\usepackage{graphicx}
\usepackage{booktabs}
\usepackage{array}
\usepackage{caption}
\usepackage{subcaption}    
\usepackage{float}         

\usepackage{natbib}        

\usepackage[hidelinks]{hyperref}

\usepackage{microtype}     

\title{A Flexible ROI Analysis Tool for Spatially Consistent Vegetation Index Extraction from Turfgrass Imagery}

\author[1]{Kibon Ku}
\author[2,3,4]{Thai Thanh Tuan}
\author[5]{Jinhyun Ahn\thanks{Corresponding author:
           \href{mailto:jha@jejunu.ac.kr}{jha@jejunu.ac.kr}}}

\affil[1]{Department of Electrical and Computer Engineering,
          Iowa State University, Ames, IA~50011, USA;
          \href{mailto:kibona9@iastate.edu}{kibona9@iastate.edu}}
\affil[2]{Department of Plant Resources and Environment,
          Jeju National University, Jeju~63243, Republic of Korea;
          \href{mailto:thaithanhtuan@jejunu.ac.kr}{thaithanhtuan@jejunu.ac.kr}}

\affil[3]{Multimedia Communications Laboratory, 
    University of Information Technology, Ho Chi Minh city 70000, Vietnam}
\affil[4]{Vietnam National University, Ho Chi Minh city 70000, Vietnam}

\affil[5]{Department of Management Information Systems,
          Jeju National University, Jeju~63243, Republic of Korea;
          \href{mailto:jha@jejunu.ac.kr}{jha@jejunu.ac.kr}}

\date{}

\begin{document}
\maketitle

\section*{Plain Language Summary}

\noindent\textbf{Subject Overview.}
Researchers capture multispectral imagery of turfgrass to
calculate vegetation indices that quantify turfgrass
health over time.

\noindent\textbf{Research Purpose.} 
Following the same plots across repeated images is difficult when
camera position shifts between sessions, and manually re-marking plots
is slow and inconsistent. We developed open-source software in which
users draw rotatable boxes over plots, divide them into equal sections,
and save their exact image locations.

\noindent\textbf{Key Findings.} 
The saved locations can be reused for every vegetation index calculated
from the same image, ensuring that all indices are measured over
identical areas.

\noindent\textbf{Key Takeaways.}
The software supports reproducible plot-level measurement in small
experiments where camera position varies, without training in mapping
software.

\section*{Abstract}

Time-series multispectral imaging is widely used to assess turfgrass
condition; however, extracting plot-level vegetation index (VI) values
remains labor-intensive when camera positions vary between
acquisitions and automated plot tracking is unreliable. We developed
an open-source software tool comprising (i)~a graphical user
interface (GUI) for the interactive placement of orientation-adjustable
regions of interest (ROIs) with $N$-way subdivision to match plot
geometry and (ii)~a batch-processing script that computes plot-level
mean VI values from single-band floating-point rasters and exports
measurements and ROI coordinates in JSON format for reuse. Saved
coordinates can be reapplied to measure NDVI, NDRE, and SAVI from
identical pixel regions within a single time point without re-delineating
ROIs. In a demonstration using rail-mounted multispectral turfgrass imagery
(18 images), plot-level mean values were extracted in a single
session, with coordinates reused directly for NDRE and SAVI rasters. The system reads non-georeferenced pixel-coordinate rasters
natively, does not require GIS expertise, and supports reproducible VI
quantification where static sensor deployment is impractical.

\section*{Abbreviations}
CRS, coordinate reference system;
GUI, graphical user interface;
GIS, geographic information system;
JSON, JavaScript Object Notation;
NDRE, normalized difference red edge index;
NDVI, normalized difference vegetation index;
NIR, near-infrared;
ROI, region of interest;
SAVI, soil-adjusted vegetation index;
UAV, unmanned aerial vehicle;
VI, vegetation index.

\section{Introduction}

Multispectral imaging from unmanned aerial vehicles (UAVs) and
ground-based platforms is increasingly used to monitor turfgrass
canopy condition for both research and agronomic
management \citep{zhang2019uav,ku2023zoysia,mcloughlin2025turfgrass}.
Vegetation indices (VIs) derived from these images---such as the
Normalized Difference Vegetation Index
(NDVI; \citealt{rouse1974erts}), the Normalized Difference Red Edge
Index (NDRE; \citealt{thompson2019ndre}), and the Soil-Adjusted
Vegetation Index (SAVI; \citealt{huete1988savi})---provide
quantitative summaries of canopy vigor and stress that can be tracked
longitudinally across experimental time points.  While the
band-arithmetic operations needed to compute individual VIs are
straightforward, converting a large time-series collection of VI
images into database-ready, plot-level measurements is more demanding:
the same plot boundaries must be identified consistently in every
image, even when sensor position or viewing geometry shifts between
acquisitions.

Geographic information system (GIS) platforms such as QGIS and
ArcGIS Pro support raster visualization and region-based zonal
statistics and are widely used for plot-level
analysis \citep{qgis,arcgispro,wilber2022arcgis}.  However, these
platforms are designed primarily for georeferenced spatial data with
assigned coordinate reference systems (CRS).  When VI imagery is not
georeferenced---as is common with raw drone outputs and many bench-top
multispectral sensors---applying GIS zonal statistics requires
intermediate georeferencing or CRS-assignment steps that add unnecessary processing complexity.  To our knowledge, neither QGIS nor
ArcGIS Pro provides a native, interactive operation for defining a
rotated parent rectangle aligned with an oblique plot row and
subdividing it into $N$ equal-width child regions in a single step.
When camera position shifts between time points, users must
re-delineate plot boundaries manually for each acquisition, which is both labor-intensive and highly prone to inter-session inconsistency \citep{lottes2020easympe}. If multiple VIs are
processed separately, slight positional differences between
independently drawn regions can introduce inter-index spatial
mismatches that complicate comparative analysis \citep{li2020uavreview}.
Although sophisticated high-throughput phenotyping pipelines
exist \citep{sun2023phenotyping,gano2024uavreview}, they typically
require substantial computational infrastructure and GIS expertise,
leaving a gap for lightweight, interactive tools suited to
small-scale experimental plots in practical field settings.

We therefore developed a Python-based open-source software that standardizes plot-region definition for
non-georeferenced VI imagery and automates extraction of plot-level VI
summaries.  The first component is an interactive GUI, built with
PyQt5 and OpenCV \citep{pyqt5,opencv}, that supports
plot-geometry-aware operations: adjustable rotatable boxes and $N$-way
subdivision aligned to bed geometry.  The second component applies
saved JSON-encoded ROI coordinates \citep{jsonstd} in batch across
multiple VI images from the same time point, ensuring pixel-level
spatial identity across indices. The objective of this study was to describe the design of this system and to demonstrate, on a rail-mounted multispectral dataset, how it supports plot-level VI extraction from identical ROIs across indices
without requiring georeferenced image data.

\section{Materials and Methods}

\subsection{Study Scenario and Input Data}

This study targets time-series turfgrass monitoring scenarios in which
repeated UAV or sensor-based acquisitions are collected at multiple
time points but imaging geometry is not perfectly reproducible between images or sessions. 
Images were acquired with a Phantom 4 Multispectral (WM336, DJI) unmanned aerial vehicle (UAV), equipped with six 1/2.9-inch CMOS sensors. The UAV was mounted on a ladder-rail system and moved along the table holding the turfgrass pots.
Because the UAV was translated along the rail during capture, the sensor
position differed between images. Input data consist of single-band, floating-point TIFF rasters (\texttt{*.tif}/\texttt{*.tiff}) representing precomputed vegetation indices---specifically NDVI, NDRE, and SAVI---derived from multispectral imagery prior to use of the proposed system \citep{rouse1974erts,thompson2019ndre,huete1988savi}.
Images are non-georeferenced (no CRS metadata assigned) and are
referenced in pixel coordinates only.  Each time point comprises
multiple VI rasters covering the same set of turfgrass experimental
plots.  As illustrated in Figure~\ref{fig:workflow}, the
proposed workflow operates on these rasters to (i)~define
plot-aligned measurement regions and (ii)~extract plot-level summary
statistics while maintaining consistent spatial sampling across all
VIs within a time point.

The demonstration dataset comprised 18 NDVI raster images acquired on the same date.
The number of parent regions of interest (ROIs) per image and the subdivision count $N$ are
user-defined and depend on the experimental layout. In this demonstration, each parent ROI was subdivided
into $N = 4$ replicates. The JSON-encoded ROI coordinates of each image were subsequently
applied to the corresponding NDRE and SAVI rasters from the same acquisition.

\subsection{System Architecture}

\begin{figure}[H]
    \centering
    \includegraphics[width=\linewidth]{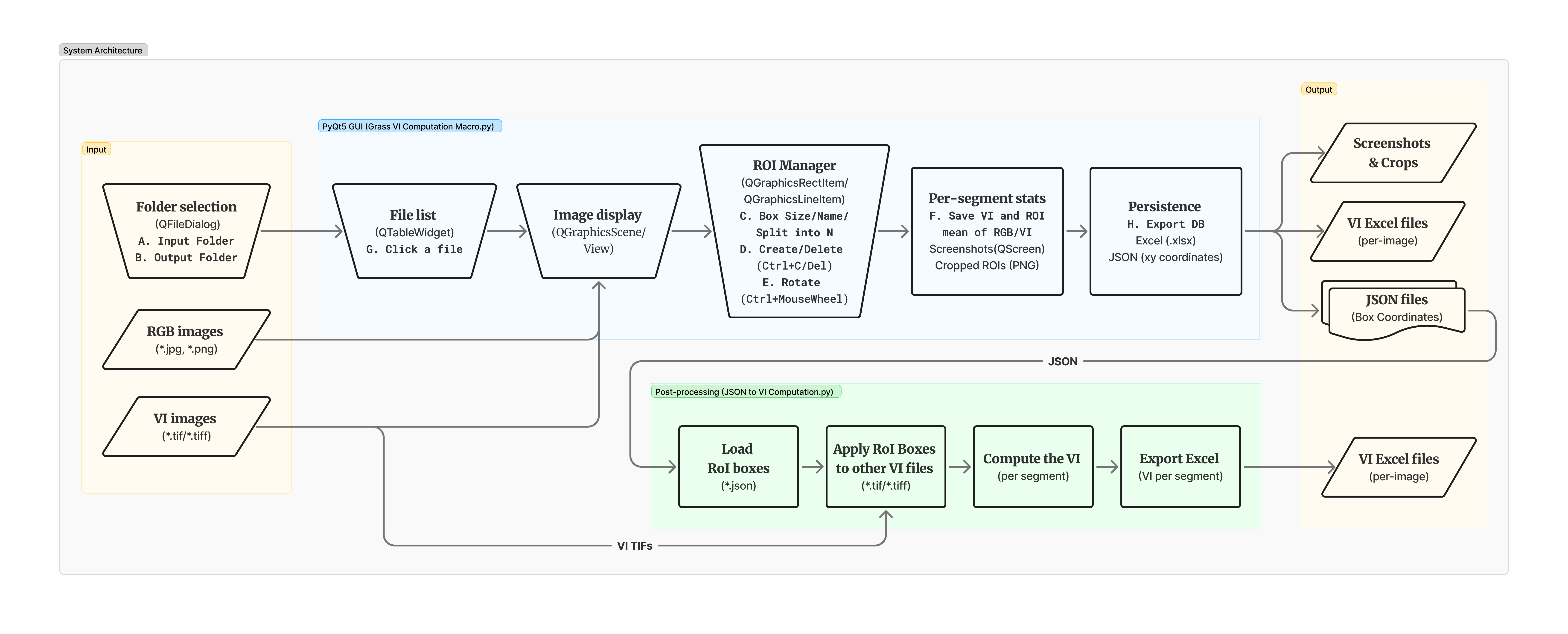}
    \caption{Two-part workflow architecture of the proposed ROI
    analysis software.}
    \label{fig:workflow}
\end{figure}

The system consists of two Python scripts.  As shown in
Figure~\ref{fig:workflow}, the upper workflow handles interactive
region definition via the GUI (\texttt{Grass VI Computation Macro.py}),
per-image VI extraction, and generation of screenshots, per-image
Excel files, and JSON coordinate files.  The lower workflow
(\texttt{JSON to VI Computation.py}) loads the saved JSON coordinates
and applies them to additional VI rasters from the same time point,
producing per-segment VI Excel files without repeated manual ROI
placement.  The GUI was implemented with PyQt5~(v5.15.4) for
interface components and OpenCV~(v4.5.2) for image rendering and
geometric operations \citep{pyqt5,opencv}.  All source code is
freely available at
\url{https://github.com/kibonku/VI-ROI-Computation-Macro}.

\subsubsection{Software Dependencies and Command-Line Interface}
\label{sec:cli}

The system requires Python~$\geq$~3.5 and four open-source packages
(PyQt5~$\geq$~5, opencv-contrib-python~$\geq$~4, XlsxWriter, and
pandas), all of which are listed in the repository
\texttt{requirements.txt} and installable via \texttt{pip}.
The batch post-processing script additionally exposes a command-line
interface accepting three flags (\texttt{-j}~JSON coordinate file,
\texttt{-i}~input raster directory, \texttt{-o}~output directory),
enabling integration into automated phenotyping pipelines without
launching the GUI.

\subsubsection{Region Definition and Parameterization}

ROIs are represented as parameterized rotated rectangles.  Each ROI
is defined by its center coordinates $(x_c,\,y_c)$ in pixel space,
width $w$ (px), height $h$ (px), and in-plane rotation angle
$\theta$ (degrees, clockwise from the image horizontal axis).  The
four corner vertices $\mathbf{v}_i$ are obtained by applying a
standard 2-D rotation matrix to the axis-aligned corner offsets and
translating to the center:

\begin{equation}
    \mathbf{v}_i = \begin{bmatrix}\cos\theta & -\sin\theta \\
                                   \sin\theta &  \cos\theta
                   \end{bmatrix}
                   \mathbf{u}_i + \begin{bmatrix}x_c\\y_c\end{bmatrix},
    \label{eq:rotation}
\end{equation}

\noindent where $\mathbf{u}_i \in \{(\pm w/2,\, \pm h/2)\}$ are the
four unrotated corner offsets.  This parameterization allows
interactive adjustment of region size and orientation to accommodate
varying camera viewpoints without requiring georeferenced input data.

To support plot-scale sampling, the GUI provides an $N$-way
subdivision operation that splits a user-defined parent rectangle
into $N$ equal-width child ROIs along the long axis (bed direction).
Each child ROI inherits $\theta$ and $h$, has width $w/N$, and is centered at
\begin{equation}
  \mathbf{c}_k = \begin{bmatrix}x_c\\y_c\end{bmatrix}
  + \left(k-\tfrac{N-1}{2}\right)\frac{w}{N}
    \begin{bmatrix}\cos\theta\\\sin\theta\end{bmatrix},
  \quad k=0,\dots,N-1.
  \label{eq:subdivision}
\end{equation}
This allows the user to define an entire plot row once and obtain a
consistent, geometry-matched set of per-replicate sampling regions
in a single operation.

\subsubsection{Vegetation-Index Value Extraction}

The mean VI value for each ROI is calculated by averaging the pixel values contained within the boundaries of the rotated rectangle.  The ROI is rasterized as a
binary mask in image coordinate space: a pixel is included if its
center falls inside the convex polygon defined by the four vertices
$\mathbf{v}_i$ (Eq.~\ref{eq:rotation}).  Pixels carrying missing or
invalid values (IEEE \texttt{NaN} or a designated no-data sentinel, if
present in the input raster) are excluded from the mean.  VI rasters
are read with the \texttt{cv2.IMREAD\_UNCHANGED} flag to preserve
32-bit or 64-bit floating-point precision.  For each processed image,
the system outputs a table of ROI identifiers and their corresponding
mean VI values.

An optional linear normalization step rescales VI values to $[0, 1]$
by applying $\text{VI}_{\text{norm}} = (\text{VI} -
\text{VI}_{\min}) / (\text{VI}_{\max} - \text{VI}_{\min})$, where
$\text{VI}_{\min}$ and $\text{VI}_{\max}$ are user-specified clip
bounds (defaulting to the theoretical index range).  This step is
optional and does not affect the spatial identity of the sampled
regions; users working with raw index values may disable it.

\subsubsection{Coordinate Export and Reuse Across Indices}
\label{sec:coords}

Ensuring pixel-level spatial consistency across multiple VIs within a single time point was a primary design objective.  After region placement, the GUI serializes each ROI's identifier, four corner vertices
(top-left, top-right, bottom-right, and bottom-left pixel
coordinates), and parameterized fields $(x_c, y_c, w, h, \theta)$ to
a structured JSON file \citep{jsonstd}.  The batch post-processing
script loads this JSON file and applies the stored coordinates to any
VI raster sharing the same pixel dimensions, requiring only the
substitution of the input raster path.  This mechanism ensures that
NDVI, NDRE, SAVI, and any additional VI rasters from the same
acquisition are always measured over exactly the same set of pixels,
eliminating inter-index spatial discrepancies that would otherwise
arise from independently re-drawn regions.

\subsection{Graphical User Interface}

\begin{figure}[H]
    \centering
    \includegraphics[width=\linewidth]{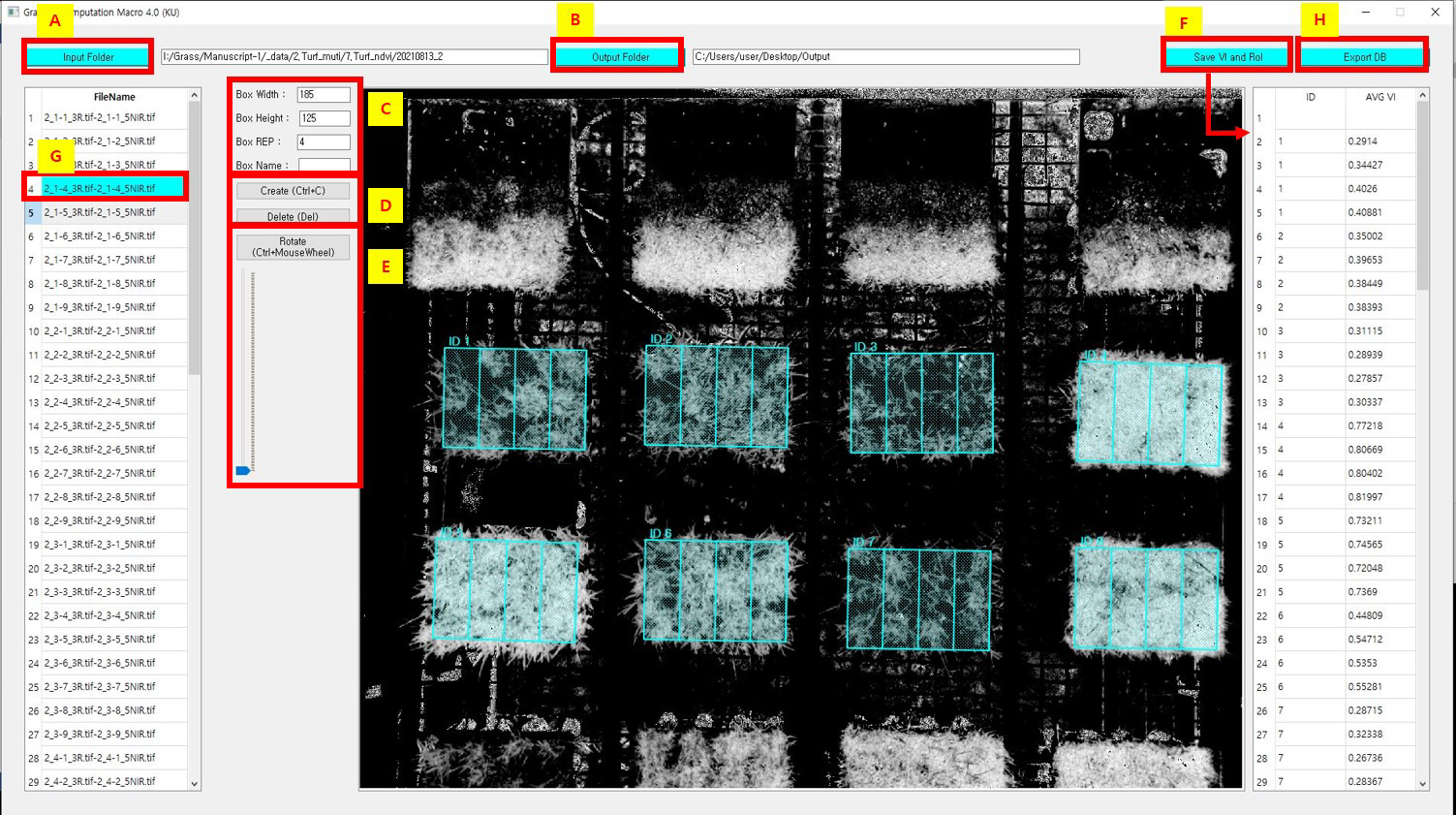}
    \caption{Graphical user interface (GUI) of the Grass VI
Computation Software (version~4.0). The screenshot is from a separate
session and is shown to illustrate the interface.}
    \label{fig:gui}
\end{figure}

The GUI is shown in Figure~\ref{fig:gui}, with labeled
components A--H.  The central canvas displays the currently loaded
VI raster with overlaid cyan ROI boxes; each parent box is
subdivided into $N$ equal-width child ROIs labeled with their
identifiers.  The AVG~VI table on the right panel displays live
per-ROI mean values as each image is processed, allowing the
operator to verify region placement before saving.

\subsubsection{Input and Output Path Configuration}
\label{sec:io}

The user selects an input directory containing the target VI rasters
via the \emph{Input Folder} button (Figure~\ref{fig:gui}A), after
which all image files in that directory are enumerated in the
\emph{FileName} list (Figure~\ref{fig:gui}G).  An output directory
for processed results is selected via the \emph{Output Folder} button
(Figure~\ref{fig:gui}B).  The software accepts single-band TIFF
rasters (\texttt{*.tif}, \texttt{*.tiff}) as VI inputs and RGB images
(\texttt{*.jpg}, \texttt{*.png}) for visual inspection.

\subsubsection{ROI Parameter Configuration and Interactive Placement}
\label{sec:roi_placement}

ROI geometry is configured via four fields (Figure~\ref{fig:gui}C):
\emph{Box Width} ($w$, pixels), \emph{Box Height} ($h$, pixels),
\emph{Box REP} ($N$, the subdivision count), and \emph{Box Name}
(user-defined identifier prefix).  Default values are $w = 185$~px,
$h = 125$~px, and $N = 4$.  Each created ROI receives a unique
identifier formed as \texttt{\{BoxName\}\{index\}} (e.g.,
\texttt{test1}, \texttt{test2}).

Clicking \emph{Create} (or pressing Ctrl\,+\,C) places a new ROI at
the image center; the ROI can then be dragged to the target plot
location (Figure~\ref{fig:gui}D).  Fine rotation to align the box
with an oblique plot row is performed via the \emph{Rotate} slider or
Ctrl\,+\,MouseWheel, with a resolution of 1$^\circ$ per scroll
increment and a maximum range of $\pm$180$^\circ$
(Figure~\ref{fig:gui}E).  Unwanted ROIs are removed with the
\emph{Delete} button or the Del key, and remaining ROI indices are
renumbered automatically.

\subsubsection{VI Computation, Mask Export, and Data Preview}
\label{sec:compute}

Once ROIs are positioned on an image, clicking \emph{Save VI and ROI}
(Figure~\ref{fig:gui}F) triggers three outputs: (i)~a screenshot of
the current canvas with overlaid ROI boxes saved to the output
directory for visual audit; (ii)~binary mask images containing only
the ROI footprints, which can be used for downstream texture analysis
or quality control; and (iii)~per-ROI mean VI values appended to the
in-memory data table.  The \emph{AVG VI} table in the right-hand side of 
Figure~\ref{fig:gui} displays live per-ROI mean values, allowing
the user to verify region placement before proceeding to the next
image.  Processed files are highlighted in blue in the file list
(Figure~\ref{fig:gui}G) to track progress across a batch.

\subsubsection{Batch Data Export}
\label{sec:export}

Upon completing all images in a session, clicking \emph{Export DB}
(Figure~\ref{fig:gui}H) writes two files to the output directory:
(i)~an Excel workbook (\texttt{.xlsx}) containing all ROI identifiers
and their per-image mean VI values, structured with one column per
image file; and (ii)~a JSON file encoding the ROI coordinate metadata
(identifier, vertices, and parameterized rectangle fields) for all
ROIs defined during the session.  The JSON file is subsequently used
by the batch post-processing script to apply the same ROI coordinates
to additional VI rasters from the same time point, as described in
Section~\ref{sec:coords}.

\section{Results and Discussion}

\subsection{ROI Placement and Subdivision on NDVI Imagery}

\begin{figure}[H]
    \centering
    \begin{subfigure}[t]{0.49\linewidth}
        \includegraphics[width=\linewidth]{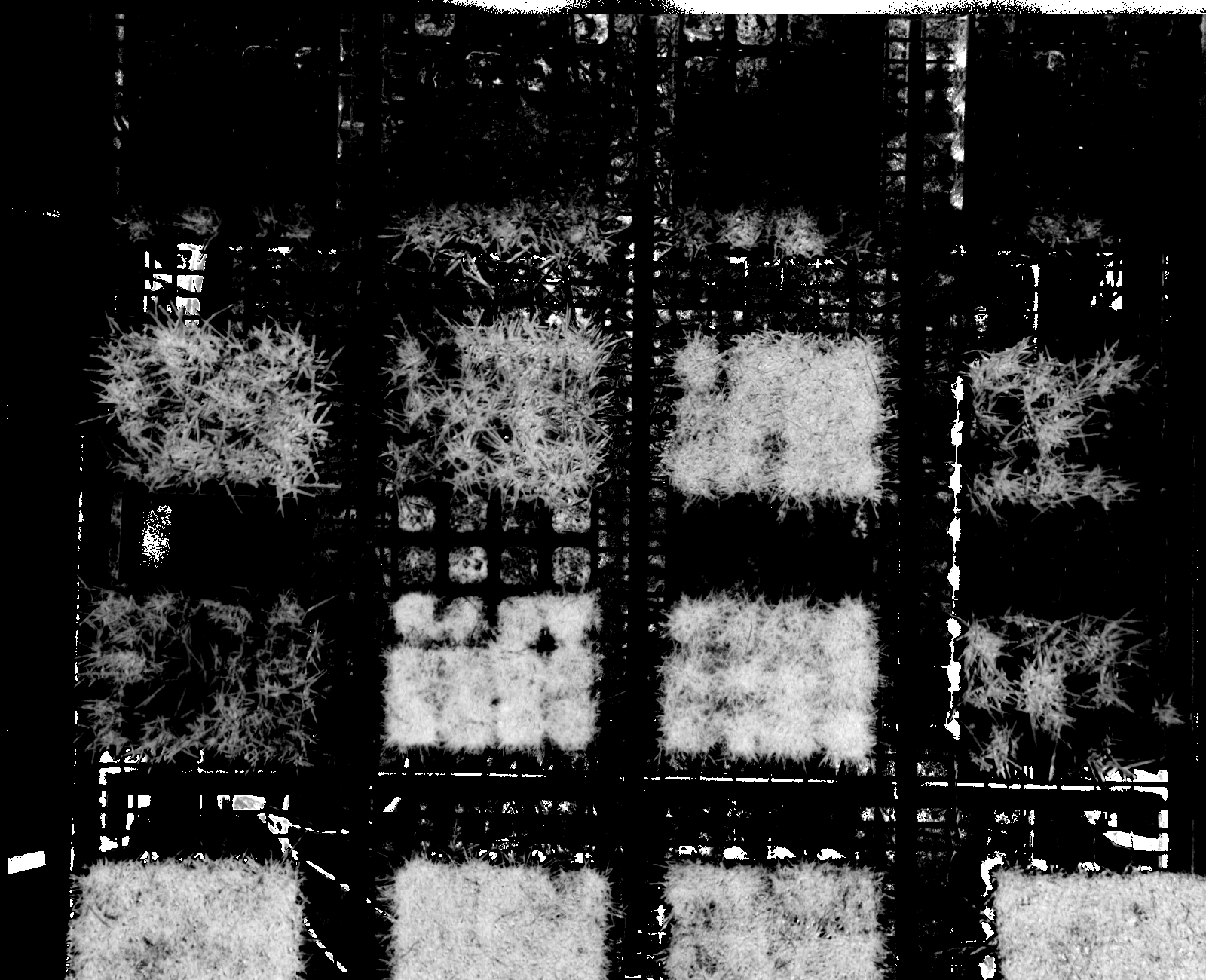}
        \caption{Raw NDVI raster, sample group~1.}
        \label{fig:raw1}
    \end{subfigure}
    \hfill
    \begin{subfigure}[t]{0.49\linewidth}
        \includegraphics[width=\linewidth]{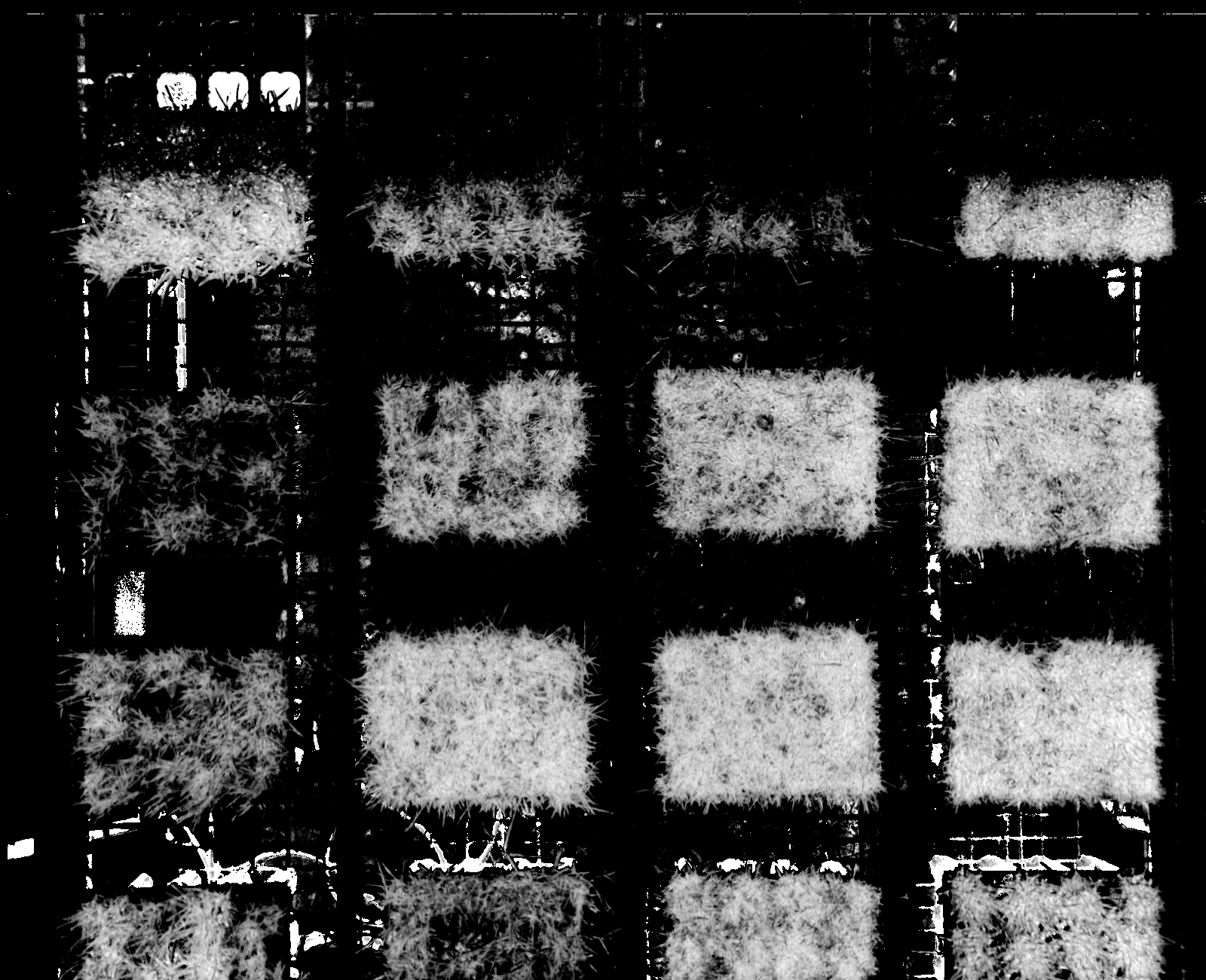}
        \caption{Raw NDVI raster, sample group~2.}
        \label{fig:raw2}
    \end{subfigure}

    \vspace{0.5em}

    \begin{subfigure}[t]{0.49\linewidth}
        \includegraphics[width=\linewidth]{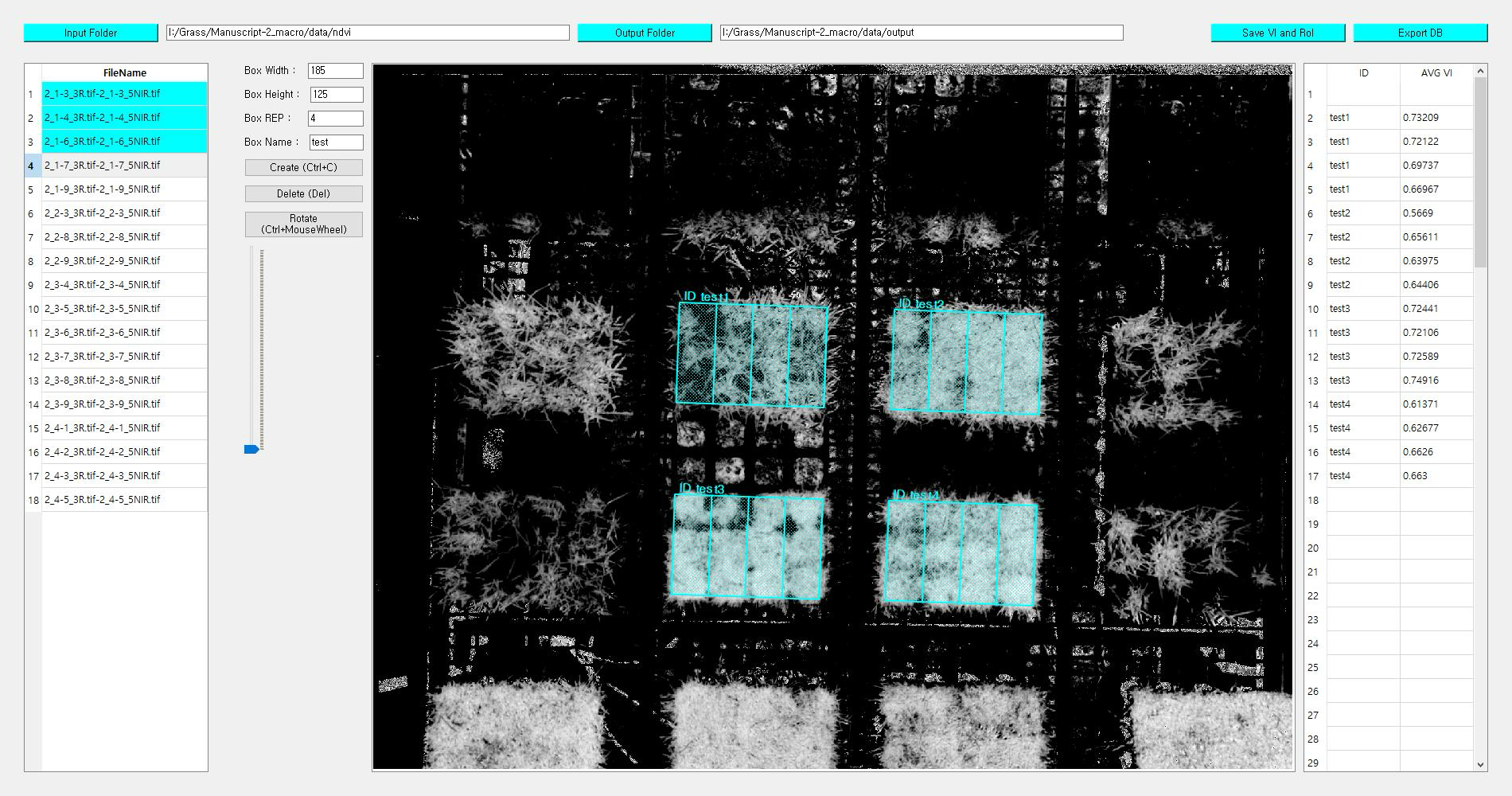}
        \caption{ROI placement on sample group~1.}
        \label{fig:roi1}
    \end{subfigure}
    \hfill
    \begin{subfigure}[t]{0.49\linewidth}
        \includegraphics[width=\linewidth]{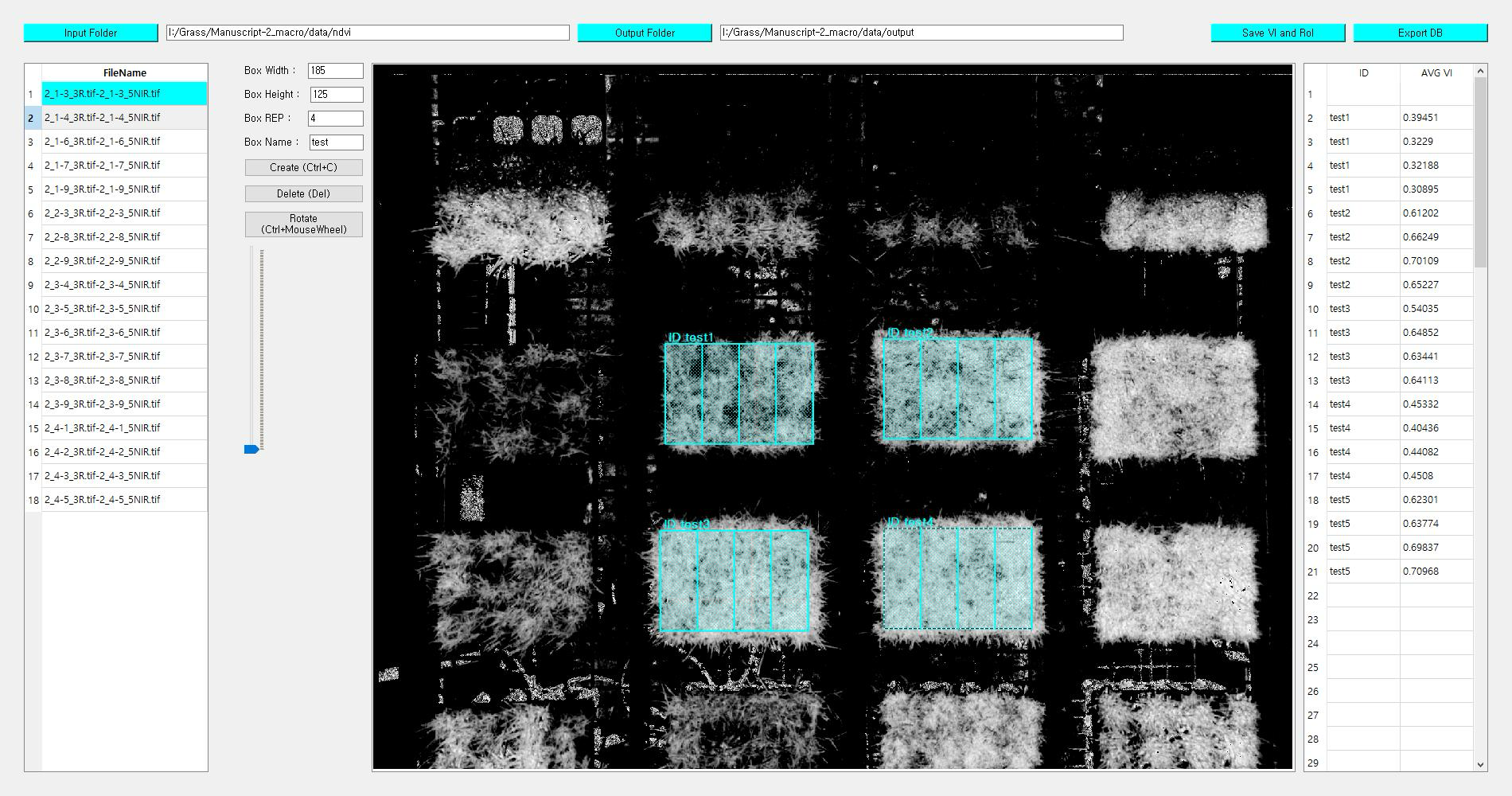}
        \caption{ROI placement on sample group~2.}
        \label{fig:roi2}
    \end{subfigure}
    \caption{Raw NDVI rasters and corresponding ROI placements for
    two turfgrass sample groups.}
    \label{fig:roi_placement}
\end{figure}

The GUI was applied to 18 NDVI raster images representing two turfgrass sample groups acquired on the same date. Images were collected using a multispectral sensor mounted on a ladder-rail system pulled along the table on which the turfgrass pots were arranged.  Figure~\ref{fig:roi_placement} shows the raw NDVI rasters (panels (a) and (b)) and the corresponding ROI placements (panels (c)
and (d)) for the two sample groups.  Because the pots were physically
arranged at an oblique angle relative to the sensor path, the ROI
boxes required rotation to align precisely with each pot boundary.
The rotation and $N$-way subdivision controls allowed the operator to fit
each ROI to the actual pot geometry. Once saved, the JSON coordinate file
of each image was applied to its corresponding NDRE and SAVI rasters from
the same acquisition without further manual input.

Each parent ROI was subdivided into $N = 4$ equal-width child
regions, yielding uniquely identified subregions for each parent ROI
(e.g., \texttt{test1}--\texttt{test4}, each with four replicates). Comparing panels (a) and (c) in Figure~\ref{fig:roi_placement},
the oblique orientation of the pots in the raw raster is clearly
reflected in the rotated ROI boxes, indicating that the rotation function visually aligns the sampling regions to the physical pot
boundaries regardless of pot arrangement.  The same behavior is
demonstrated for sample group~2 in panels (b) and (d).

The exported mean NDVI values are summarized in
Table~\ref{tab:ndvi} and ranged from approximately 0.26 to 0.77, indicating that the extracted values can resolve differences in
canopy vigor among turfgrass pots \citep{ku2023zoysia}.  The live AVG~VI table in
the GUI enabled the operator to identify implausible values arising
from non-vegetated border areas and adjust ROI placement before
export.

\begin{table}[H]
\centering
\caption{Mean NDVI values extracted by the GUI for nine of the 18
images, shown for parent ROIs \texttt{test1}--\texttt{test4}.}
\label{tab:ndvi}
\small
\setlength{\tabcolsep}{4pt}
\begin{tabular}{lcccc}
\toprule
\textbf{Image filename}$^{a}$ & \textbf{test1} & \textbf{test2} & \textbf{test3} & \textbf{test4} \\
\midrule
2\_1-3\_3R.tif & 0.3945 & 0.6120 & 0.5404 & 0.4533 \\
2\_1-4\_3R.tif & 0.4915 & 0.6056 & 0.7092 & 0.7363 \\
2\_1-6\_3R.tif & 0.7321 & 0.5669 & 0.7244 & 0.6138 \\
2\_1-7\_3R.tif & 0.3167 & 0.6027 & 0.6906 & 0.6303 \\
2\_1-9\_3R.tif & 0.5947 & 0.5556 & 0.5226 & 0.5182 \\
2\_2-3\_3R.tif & 0.2939 & 0.4757 & 0.3054 & 0.4169 \\
2\_2-8\_3R.tif & 0.6792 & 0.3903 & 0.4730 & 0.5991 \\
2\_2-9\_3R.tif & 0.2712 & 0.4228 & 0.2630 & 0.4556 \\
2\_3-4\_3R.tif & 0.3538 & 0.4882 & 0.7738 & 0.2759 \\
\bottomrule
\end{tabular}

\vspace{0.3em}
{\footnotesize $^{a}$Paired NIR suffix omitted for brevity;
full filenames are recorded in the exported \texttt{.xlsx} file.
Values rounded to four decimal places.}
\end{table}

\subsection{Cross-Index Coordinate Reuse}

\begin{figure}[H]
    \centering
    \begin{subfigure}[t]{0.32\linewidth}
        \includegraphics[width=\linewidth]{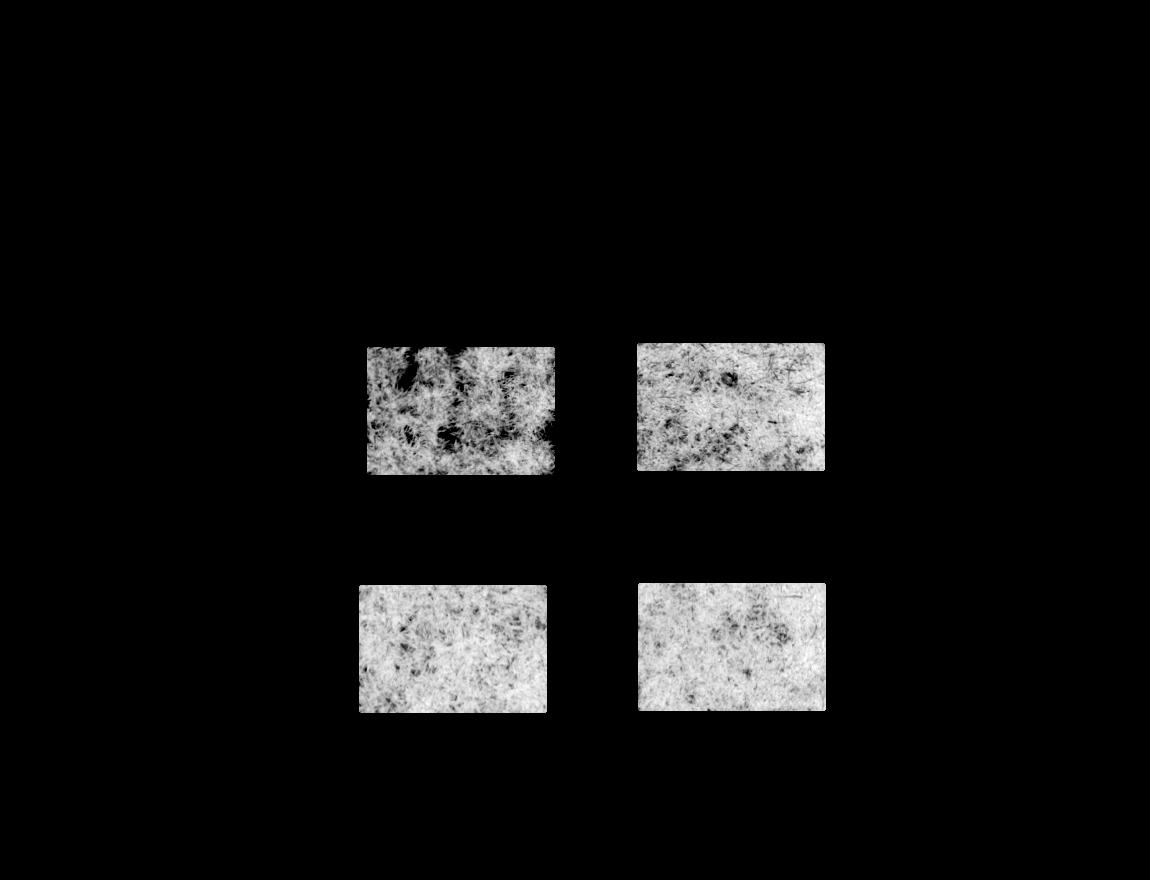}
        \caption{NDVI mask.}
        \label{fig:mask_ndvi}
    \end{subfigure}
    \hfill
    \begin{subfigure}[t]{0.32\linewidth}
        \includegraphics[width=\linewidth]{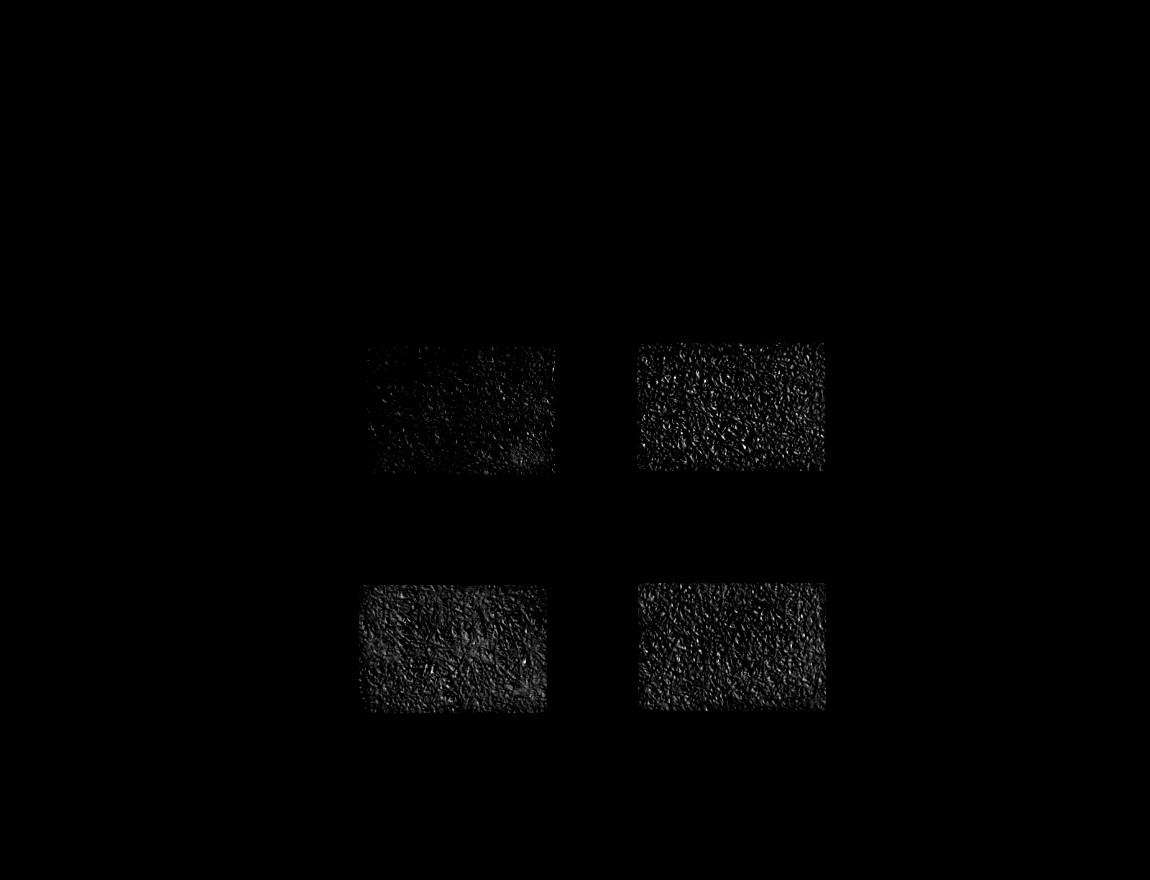}
        \caption{SAVI mask.}
        \label{fig:mask_savi}
    \end{subfigure}
    \hfill
    \begin{subfigure}[t]{0.32\linewidth}
        \includegraphics[width=\linewidth]{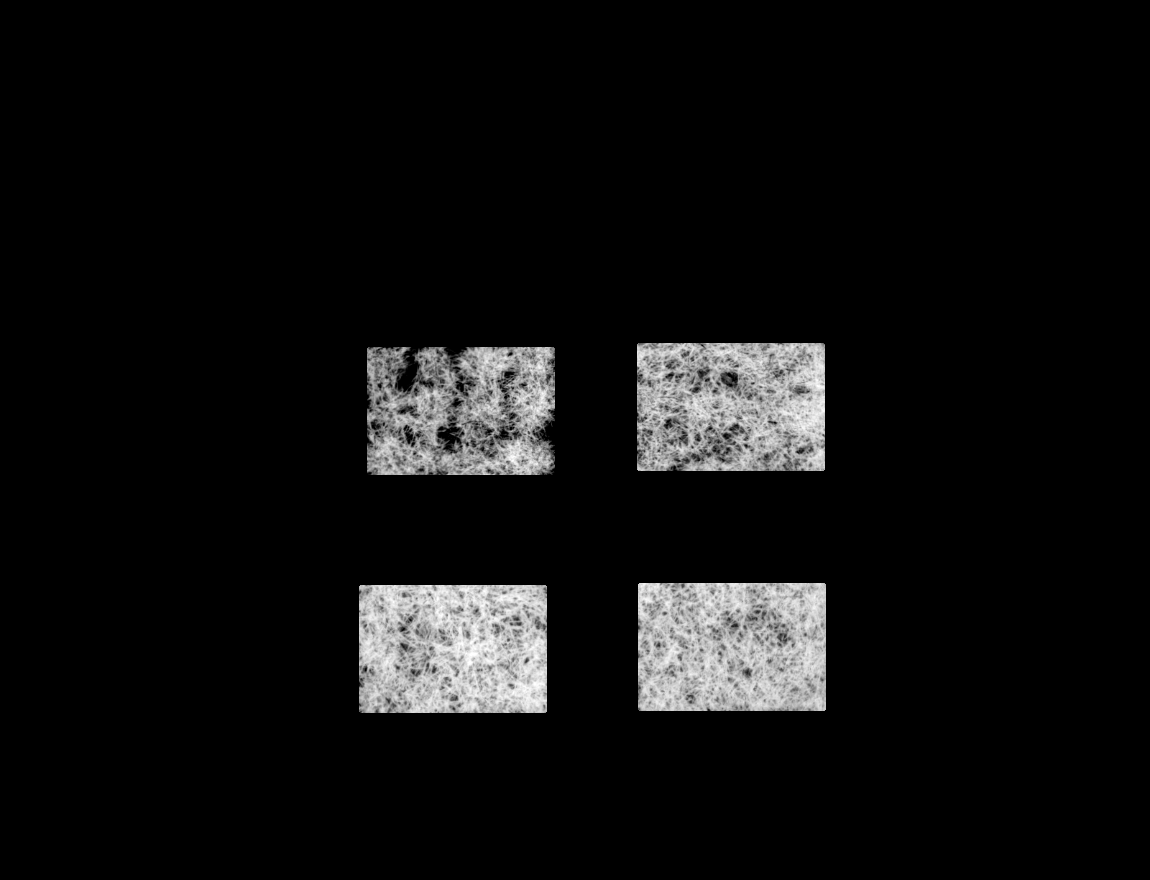}
        \caption{NDRE mask.}
        \label{fig:mask_ndre}
    \end{subfigure}
    \caption{Binary ROI mask images exported for three vegetation
    indices from the same JSON coordinate file.}
    \label{fig:masks}
\end{figure}

To demonstrate cross-index spatial consistency, the JSON coordinate
file produced from the NDVI session was supplied as input to the
batch post-processing script along with the corresponding NDRE and
SAVI rasters from the same acquisition.  No ROI re-delineation was
performed.  Figure~\ref{fig:masks} shows the binary mask images
exported for NDVI (A), SAVI (B), and NDRE (C) from the same set of
ROI coordinates.  The three mask images are pixel-identical by construction, because the
same coordinates are applied to all three rasters. This avoids the
positional differences that can arise when polygons are placed
independently for each index, which may bias inter-index comparisons,
particularly at plot boundaries where index gradients are steep
\citep{li2020uavreview}.

The ROI coordinate metadata exported to JSON is summarized in
Table~\ref{tab:json} for ROI group \texttt{test1} of the first
image.  Each row corresponds to one child ROI produced by the $N=4$
subdivision; these coordinates are loaded verbatim by the batch
script and applied to the NDRE and SAVI rasters without modification,
ensuring pixel-level spatial identity across indices.

\begin{table}[H]
\centering
\caption{Corner-vertex pixel coordinates of child ROIs for ROI
group \texttt{test1}, image \texttt{2\_1-3\_3R.tif}, exported to
\texttt{ndvi.json}.}
\label{tab:json}
\small
\setlength{\tabcolsep}{4pt}
\begin{tabular}{lcccc}
\toprule
\textbf{Child ROI} &
\textbf{TL $(x,y)$} & \textbf{TR $(x,y)$} &
\textbf{BR $(x,y)$} & \textbf{BL $(x,y)$} \\
\midrule
rep\,1 & (374, 322) & (420, 322) & (420, 447) & (374, 447) \\
rep\,2 & (420, 322) & (466, 322) & (466, 447) & (420, 447) \\
rep\,3 & (466, 322) & (513, 322) & (513, 447) & (466, 447) \\
rep\,4 & (513, 322) & (559, 322) & (559, 447) & (513, 447) \\
\bottomrule
\end{tabular}

\vspace{0.3em}
{\footnotesize
TL, top-left; TR, top-right; BR, bottom-right; BL, bottom-left.
Coordinates are integer pixel values (origin at image top-left
corner).}
\end{table}

\subsection{Comparison with GIS-Based Workflows}

General-purpose GIS platforms such as QGIS and ArcGIS Pro offer
raster zonal statistics functionality and Python-based batch
processing (PyQGIS and \texttt{arcpy}, respectively) that can in
principle automate plot-level VI extraction \citep{qgis,arcgispro,wilber2022arcgis}. However, three key operational differences emerge when processing non-georeferenced, time-series multispectral imagery.

First, zonal statistics in QGIS and ArcGIS Pro are typically performed
on rasters that carry a valid CRS, so non-georeferenced rasters may
need a CRS assigned beforehand. Assigning a CRS to such rasters adds a preparatory step for every image. In contrast, the proposed software tool reads pixel-coordinate rasters directly, without requiring any CRS assignment.

Second, to our knowledge, neither QGIS nor ArcGIS Pro provides a native GUI operation to place a rotated parent rectangle over an oblique plot row and 
subdivide it into $N$ child regions in a single interactive step.
Achieving this in QGIS requires custom scripting (via PyQGIS or the
graphical modeler), and resulting layout is not interactively adjustable
on the raster canvas in real time.

Third, the JSON-based coordinate reuse mechanism ensures that all
VIs measured at a given time point utilize pixel-identical ROIs. In a conventional
GIS workflow, each VI raster is typically processed through a
separate zonal-statistics operation, requiring the same vector layer to be re-imported or re-linked for each index. While scripting can automate this process, it requires specialized GIS scripting skills that are less common in plant science laboratories.

While GIS platforms remain the appropriate choice for large-scale, georeferenced spatial analysis, the proposed software tool targets a more focused use case: extracting plot-level VI data from non-georeferenced rasters when camera positions vary between sessions and GIS expertise is limited.

\section{Conclusion}

We developed an open-source software tool to address a practical bottleneck in time-series turfgrass phenotyping: the inability to reliably re-identify the same plot regions across multispectral acquisitions when camera position is not reproducible. The interactive GUI enables users to place orientation-adjustable, $N$-way subdividable ROIs directly on non-georeferenced VI rasters, avoiding the CRS-assignment step that GIS platforms typically require. The JSON-based coordinate export mechanism ensures pixel-level spatial identity across all VI layers derived from the same acquisition, preventing the inter-index spatial discrepancies that can arise when ROIs are drawn independently for each index. A demonstration on rail-mounted multispectral turfgrass imagery confirmed that plot-level VI values could be extracted in a single interactive session, with the same ROI coordinates reapplied to NDRE and SAVI rasters without additional manual input, yielding pixel-identical spatial footprints across all indices.

Five limitations of the current system are noted.
First, no quantitative comparison of processing time or accuracy against
GIS-based workflows or fully manual delineation was performed;
such benchmarking is needed to quantify the efficiency gains.
ROI placement also remains a manual step, introducing operator subjectivity and limiting throughput for large image sets; template-matching or feature-based image registration \citep{li2020uavreview} could in future provide an initial automated ROI estimate for the user to confirm or adjust. The demonstration relied on a single sensor and experimental setup, so validation across other sensors, spatial resolutions, and crop species is needed before broader applicability can be claimed. Additionally, the system currently reports only per-ROI mean VI values; incorporating variance, coefficient of variation, and percentile summaries would extend its utility for canopy heterogeneity analysis. Distributing the software as a self-contained executable would further reduce the installation burden for researchers without Python experience.

Despite these limitations, the proposed system fills a practical gap for plant scientists who require reproducible, spatially consistent VI extraction from non-georeferenced imagery without GIS expertise. The source code is openly available, and the lightweight architecture makes it readily adaptable to other plot-scale phenotyping applications where static sensor deployment is impractical.

\section*{Author Contributions}

Kibon Ku: Conceptualization, Data curation, Formal analysis,
Investigation, Resources, Software, Validation, Visualization,
Writing -- Original Draft, Writing -- Review \& Editing.
Thai Thanh Tuan: Data curation, Software, Validation.
Jinhyun Ahn: Funding acquisition, Project administration,
Supervision, Writing -- Review \& Editing.

\section*{Acknowledgments}
This work was supported by the National Research Foundation of Korea (NRF) grant funded by the Korean government (MSIT) (RS-2024-00348897).

\section*{Conflict of Interest}
The authors declare no conflict of interest.

\section*{Data Availability}
\begin{sloppypar}
The source code for both software components
\texttt{Grass\_VI\_Computation\_Macro.py} and
\texttt{JSON~to~VI~Computation.py}) is openly available at
\url{https://github.com/kibonku/VI-ROI-Computation-Macro}.
The repository includes a \texttt{requirements.txt} file specifying
all Python dependencies (Python~$\geq$~3.5, PyQt5~$\geq$~5,
opencv-contrib-python~$\geq$~4, XlsxWriter, and pandas), enabling
environment setup with a single \texttt{pip install} command.
The demonstration dataset is available from the corresponding author
upon reasonable request.
\end{sloppypar}

\section*{ORCID}
Kibon Ku: \url{https://orcid.org/0009-0006-5822-7103}\\
Thai Thanh Tuan: \url{https://orcid.org/0000-0003-2748-0529}\\
Jinhyun Ahn: \url{https://orcid.org/0000-0002-2331-004X}

\bibliographystyle{plainnat}
\bibliography{references}

\begin{thebibliography}{16}
\providecommand{\natexlab}[1]{#1}
\providecommand{\url}[1]{\texttt{#1}}
\expandafter\ifx\csname urlstyle\endcsname\relax
  \providecommand{\doi}[1]{doi: #1}\else
  \providecommand{\doi}{doi: \begingroup \urlstyle{rm}\Url}\fi

\bibitem[Bradski(2000)]{opencv}
Gary Bradski.
\newblock The {OpenCV} library (opencv-contrib-python version~4.5.2.54).
\newblock Dr.\ Dobb's Journal of Software Tools, 2000.
\newblock URL \url{https://opencv.org}.
\newblock Available at \url{https://opencv.org}.

\bibitem[Bray(2017)]{jsonstd}
Tim Bray.
\newblock The {JavaScript} object notation ({JSON}) data interchange format.
\newblock RFC~8259, Internet Engineering Task Force ({IETF}), 2017.
\newblock URL \url{https://www.rfc-editor.org/rfc/rfc8259}.

\bibitem[{Esri Inc.}(2022)]{arcgispro}
{Esri Inc.}
\newblock {ArcGIS Pro}, version~3.0.
\newblock Software. Esri Inc., Redlands, CA, 2022.
\newblock URL \url{https://www.esri.com/en-us/arcgis/products/arcgis-pro/overview}.

\bibitem[Gano et~al.(2024)]{gano2024uavreview}
B{\'e}n{\'e}dicte Gano et~al.
\newblock Drone-based imaging sensors, techniques, and applications in plant phenotyping for crop breeding: {A} comprehensive review.
\newblock \emph{The Plant Phenome Journal}, 7\penalty0 (1):\penalty0 e20100, 2024.
\newblock \doi{10.1002/ppj2.20100}.

\bibitem[Huete(1988)]{huete1988savi}
Alfredo~R. Huete.
\newblock A soil-adjusted vegetation index ({SAVI}).
\newblock \emph{Remote Sensing of Environment}, 25\penalty0 (3):\penalty0 295--309, 1988.
\newblock \doi{10.1016/0034-4257(88)90106-X}.

\bibitem[Ku et~al.(2023)Ku, Mansoor, Han, et~al.]{ku2023zoysia}
Kibon Ku, Saqib Mansoor, Gang-Deuk Han, et~al.
\newblock Identification of new cold tolerant {Zoysia} grass species using high-resolution {RGB} and multi-spectral imaging.
\newblock \emph{Scientific Reports}, 13:\penalty0 13209, 2023.
\newblock \doi{10.1038/s41598-023-40128-2}.

\bibitem[Li et~al.(2020)Li, Xu, Zhang, Han, Bian, Li, Liu, and Jin]{li2020uavreview}
Bo~Li, Xiaming Xu, Lu~Zhang, Junwei Han, Chunsong Bian, Guangyao Li, Jianguo Liu, and Lei Jin.
\newblock Above-ground biomass estimation and yield prediction in potato by using {UAV}-based {RGB} and hyperspectral imaging.
\newblock \emph{Field Crops Research}, 248:\penalty0 107718, 2020.
\newblock \doi{10.1016/j.fcr.2019.107718}.

\bibitem[Lottes and Stachniss(2020)]{lottes2020easympe}
Philipp Lottes and Cyrill Stachniss.
\newblock Easy {MPE}: Extraction of quality microplot images for {UAV}-based high-throughput field phenotyping.
\newblock \emph{Plant Phenomics}, 2020:\penalty0 2734527, 2020.
\newblock \doi{10.34133/2020/2734527}.

\bibitem[McLoughlin et~al.(2025)McLoughlin, Schiavon, and Kaddoura]{mcloughlin2025turfgrass}
Patrick~H. McLoughlin, Marco Schiavon, and Youssef Kaddoura.
\newblock Evaluating warm-season turfgrass: Multispectral and red--green--blue index comparison.
\newblock \emph{International Turfgrass Society Research Journal}, 2025.
\newblock \doi{10.1002/its2.70022}.
\newblock Published online 14 March 2025.

\bibitem[{QGIS Development Team}(2025)]{qgis}
{QGIS Development Team}.
\newblock {QGIS} geographic information system.
\newblock Open Source Geospatial Foundation, 2025.
\newblock URL \url{https://qgis.osgeo.org}.

\bibitem[{Riverbank Computing Limited}(2021)]{pyqt5}
{Riverbank Computing Limited}.
\newblock {PyQt5}: {Python} bindings for the {Qt} application framework, version~5.15.4.
\newblock Software, 2021.
\newblock URL \url{https://www.riverbankcomputing.com/software/pyqt/}.

\bibitem[Rouse et~al.(1974)Rouse, Haas, Schell, and Deering]{rouse1974erts}
John~W. Rouse, Robert~H. Haas, John~A. Schell, and Donald~W. Deering.
\newblock Monitoring vegetation systems in the {Great Plains} with {ERTS}.
\newblock Technical Report 19740022614, NASA Goddard Space Flight Center, 1974.
\newblock URL \url{https://ntrs.nasa.gov/citations/19740022614}.
\newblock Third ERTS Symposium, NASA SP-351, Vol.~1, pp.~309--317.

\bibitem[Sun et~al.(2023)Sun, Li, Paterson, Jiang, Xu, Robertson, Snider, and Chee]{sun2023phenotyping}
Shuaishuai Sun, Changying Li, Andrew~H. Paterson, Yu~Jiang, Rui Xu, Jerry~S. Robertson, John~L. Snider, and Peng~W. Chee.
\newblock In-field high-throughput phenotyping of cotton plant height by {UAV} multispectral image.
\newblock \emph{Remote Sensing}, 15\penalty0 (4):\penalty0 1110, 2023.
\newblock \doi{10.3390/rs15041110}.

\bibitem[Thompson et~al.(2019)Thompson, Guo, Sharma, and Ritchie]{thompson2019ndre}
C.~N. Thompson, W.~X. Guo, B.~Sharma, and G.~L. Ritchie.
\newblock Using normalized difference red edge index to assess maturity in cotton.
\newblock \emph{Crop Science}, 59\penalty0 (5):\penalty0 2167--2177, 2019.
\newblock \doi{10.2135/cropsci2019.04.0227}.

\bibitem[Wilber et~al.(2022)Wilber, Czarnecki, and McCurdy]{wilber2022arcgis}
Amy~L. Wilber, Joby~M.P. Czarnecki, and James~D. McCurdy.
\newblock An {ArcGIS Pro} workflow to extract vegetation indices from aerial imagery of small-plot turfgrass research.
\newblock \emph{Crop Science}, 62\penalty0 (1):\penalty0 503--511, 2022.
\newblock \doi{10.1002/csc2.20669}.

\bibitem[Zhang et~al.(2019)Zhang, Virk, Porter, Kenworthy, Sullivan, and Schwartz]{zhang2019uav}
Jingcheng Zhang, Simerjeet Virk, Warren Porter, Kevin Kenworthy, Dennis Sullivan, and Brian Schwartz.
\newblock Applications of unmanned aerial vehicle based imagery in turfgrass field trials.
\newblock \emph{Frontiers in Plant Science}, 10:\penalty0 279, 2019.
\newblock \doi{10.3389/fpls.2019.00279}.

\end{thebibliography}

\end{document}